\documentclass[acmsmall]{acmart}
\AtBeginDocument{%
  \providecommand\BibTeX{{%
    \normalfont B\kern-0.5em{\scshape i\kern-0.25em b}\kern-0.8em\TeX}}}

\setcopyright{acmcopyright}
\copyrightyear{2021}
\acmYear{2021}
\acmDOI{10.1145/1122445.1122456}

\acmJournal{TALLIP}
\acmVolume{37}
\acmNumber{4}
\acmArticle{111}
\acmMonth{8}

\begin{document}

\title{Hierarchical Text Classification of Urdu News using Deep Neural Network}

\author{Taimoor Ahmed Javed}
\email{taimurahmed60@gmail.com}
\orcid{0000-0003-0624-2871}
\authornotemark[1]
\affiliation{%
  \institution{National University of Computer and Emerging Sciences, Islamabad}
  \streetaddress{H-11/4}
  \city{Islamabad}
  \state{Islamabad Capital Territory}
  \country{Pakistan}
  \postcode{44000}
}

\author{Waseem Shahzad}
\affiliation{%
  \institution{National University of Computer and Emerging Sciences, Islamabad}
  \streetaddress{H-11/4}
  \city{Islamabad}
  \country{Pakistan}}
\email{waseem.shahzad@nu.edu.pk}

\author{Umair Arshad}
\affiliation{%
  \institution{National University of Computer and Emerging Sciences, Islamabad}
  \country{Pakistan}}

\renewcommand{\shortauthors}{Taimoor and Waseem, et al.}

\begin{abstract}
 Digital text is increasing day by day on the internet. It is very challenging to classify a large and heterogeneous collection of data, which require improved information processing methods to organize text. To classify large size of corpus, one common approach is to use hierarchical text classification, which aims to classify textual data in a hierarchical structure. Several approaches have been proposed to tackle classification of text but most of the research has been done on English language. This paper proposes a deep learning model for hierarchical text classification of news in Urdu language - consisting of 51,325 sentences from 8 online news websites belonging to the following genres: Sports; Technology; and Entertainment. The objectives of this paper are twofold: (1) to develop a large human-annotated dataset of news in Urdu language for hierarchical text classification; and (2) to classify Urdu news hierarchically using our proposed model based on LSTM mechanism named as Hierarchical Multi-layer LSTMs (HMLSTM). Our model consists of two modules: Text Representing Layer, for obtaining text representation in which we use Word2vec embedding to transform the words to vector and Urdu Hierarchical LSTM Layer (UHLSTML) an end-to-end fully connected deep LSTMs network to perform automatic feature learning, we train one LSTM layer for each level of the class hierarchy. We have performed extensive experiments on our self created dataset named as Urdu News Dataset for Hierarchical Text Classification (UNDHTC). The result shows that our proposed method is very effective  for hierarchical text classification and it outperforms baseline methods significantly and also achieved good results as compare to deep neural model.
\end{abstract}



\maketitle

\section{Introduction}
Nowadays, electronic information extraction systems are omnipresent, from instant mobile messaging apps to automated archives having millions of data. Due to large amount of data, it has produce significant amount of challenges. However one initiative is to categorize some of this text content automatically, so that users can more easily access, interpret and modify data for patterns and knowledge generation. Organizing large amount of electronic data into categories is becoming a topic of interest for many individuals and companies. The only solution to this problem is text classification. Text classification is the process to categorize documents according to its categories. It uses a wide variety of areas of expertise which includes AI, NLP and machine learning. It uses a supervised learning based approach, in which we train a model by giving a large amount of data. Recently, several algorithms have been proposed like SVM, k-Nearest Neighbor, Naive Bayes. Results shows that these approaches are very effective in conventional text classification applications. There are many application of text classification like topic modeling, sentiment analysis, intent detection and spam detection. Generally text classification tasks have only few classes. When classification tasks have large number of classes then we will extend text classification problem because they have some specific challenges which require particular solution. There are numerous significant issues in classification which consists of similar categories which are organized into hierarchical structure. So here hierarchical classification comes into being. In hierarchical classification, classes are organized into categories and sub categories or we can say that classes are organized into class hierarchy and when we apply this to textual data it will become hierarchical text classification.\\* In recent years, hierarchical classification seems to be an active topic of research in the field of databases. Web directories like Yahoo, Wikipedia are example of hierarchical text repositories. There are different applications nowadays in which documents are organized as hierarchical structure. If we take a real world example, Hierarchical text classification is like a librarian job, who want to place a book in a shelf. Many organizations like IT companies, law firms and medical companies are taking advantage of automatic classification of documents. Therefore, it shows that hierarchical classification have a significant impact on many applications and organizations.\\*
The Urdu language has a global reach and it is Pakistan's national language. Urdu language is also being spoke in six states of India as their official language. There are more than 12 million speakers in Pakistan and over 300 million speakers all across the globe \cite{riaz2008baseline}. Previously, Urdu was ignored by researchers due to its complex composition, peculiar characteristics and absence of linguistic capital \cite{daud2017urdu}. Due to these features, Urdu text classification is much more complex than other languages. Moreover, automatic text classification systems which is developed for other languages cannot be used for Urdu language. Since last 10 years, we saw a rapid growth in Urdu text documents like journals, web databases and online news articles. Because of these considerations, it raise the interest of researchers in the Urdu language.\\*
Mostly research work is done in English language for hierarchical text classification. Recently, many researches shows that this topic is gaining a lot of interest nowadays. So research work on hierarchical classification has been started in many languages like Chinese, Arabic and Indonesian. There is a huge gap for Urdu language. Research is done on simple text classification but there is no research work on hierarchical Urdu text classification. However, many studies has been found on simple Urdu text classification. For instance, \cite{ali2009urdu} have done research on Urdu text classification by applying SVM on large urdu corpus. \cite{khan2015urdu} have applied different feature extraction techniques on Urdu language text and Decision Tree algorithm J-48 is used for classification. Most of the previous studies on automated text classification is based on supervised learning techniques such as decision trees, naive Bayes and support vector machines (SVM). Nevertheless automated classification has become more and more difficult due to a rising number of corpus sizes and categorizing a document into fields and subfields. Due to this, these machine learning algorithms tend to have less accuracy. In fact mostly studies perform performance on machine learning methods. Since, the proposed approach is based on deep neural network.\\*
Neural networks have recently been shown to be efficient in conducting endways learning of hierarchical text classification. To fill a gap of hierarchical text classification for Urdu language and overcome the limitations of the above machine learning models and to obtain improved results for the Urdu document classification, we introduce a hierarchical model based on LSTM mechanism named as Hierarchical Multi-layer LSTMs (HMLSTM). HMLSTM incorporates deep learning algorithms to facilitate both general and advanced learning at the document hierarchy level. In this study, we implement a supervised HMLSTM model on Urdu news to extract categories and sub categories and their hierarchical relationships. We first apply Text Representing Layer to achieve text representation in which we use Word2vec embedding to transform the words to vector and prepare words vocabulary. Then we apply Urdu Hierarchical LSTM Layer (UHLSTML) on our self created dataset named as Urdu News Dataset for Hierarchical Text Classification (UNDHTC)\footnote{https://github.com/TaimoorAhmedJaved/Urdu-News-Dataset-for-Hierarchical-Text-Classification}. In our experiments, we select six state of the art machine learning models and one deep neural model to evaluate the effectiveness of HMLSTM. The experiments on real world dataset shows the efficiency and predictive capacity of our model. In particular, our model exhibits improved accuracy over conventional text classification approaches.\\*
To summarize, this paper includes the following contributions:
\begin{itemize} 
\item Urdu language has insufficient resources, but it has a complex and complicated morphological script, which makes automated text processing more difficult. A big issue for the Urdu language is the lack of availability of large, standardized, accessible and cost-free text dataset. So in this study, we develop a large dataset of Urdu news documents and will make this dataset freely accessible for future study.
\item This research is conducted because hierarchical text classification method has never been applied to classify Urdu news using deep neural network. So, we have developed the HMLSTM method which can predict the categories of each level while classifying all categories in the whole hierarchical structure accurately. 
\item We conduct a thorough evaluation of our proposed method on Urdu News dataset (UNDHTC) from different domains. The experiment results show that our method can outperforms various baselines methods and deep neural model.
\end{itemize}
The rest of this paper is structured as follows. In the next segment, we will be discussing related work. In Section 3, we will discuss the background of hierarchical text classification process. We will address what form of class structures most widely used, that is either DAG structure or tree structure. We will also discuss the types of hierarchical text classification and analyse which type is better than other. In Section 4, we will discuss base line techniques. In Section 5, we will address our proposed methodology and dataset. Section 6 presents the experiments and results. Section 7 concludes the paper and also future work is discussed.
\section{Related Work}
This section of the paper discusses recent efforts and contribution on hierarchical text classification and their methods used in similar areas which have an influence in this field. A variety of methods have been proposed to enhance the quality of classification results which are discussed below.\\*
In conventional text classification, feature engineering has been used to get useful quality features which help in improving the model performance. Text classification is a well-proven way to organize a large scale document set. Many algorithms have been proposed \cite{b5} to do text classification, such as SVM, kNN, naïve bayes, decision trees. Analyses showed that in conventional text classification applications, these algorithms are very efficient. Nowadays, deep learning models are widely use in natural language processing tasks. Recently, Nabeel \textit{et al} \cite{b28} applied deep learning classification methodologies on Urdu text. For Arabic text classification \cite{b29} also applied deep neural models. Similarly, authors applied attention mechanism for Chinese text classification \cite{b30}.
Nonetheless, text classification has become very challenging these days. It is due to the growth in corpus sizes, categories and subcategories. With the increasing number of categories and subcategories of a document, it is important to not just label the document by a specialized field but also organized in its overall category and subcategory. So we use hierarchical classification.\par
Deep neural models have recently gained popularity in the classification of text by means of their communication capacity and low requirements for features. But deep neural models based on extensive training data and needs people to provide lots of labelled data and in many cases when we have to label data, we have to find domain expert person to do this task which can be too costly. Another issue in deep neural models is that in the hierarchical text classifications it cannot determine appropriate levels in the class hierarchy. There is no automatic way to determine best level in class hierarchy. So recently Meng \textit{et al} \cite{b6} proposed a weakly supervised neural method for hierarchical text classification which address above issues. This approach is based on deep neural network which requires only small amount of labelled data provided by the users. It doesn’t require large amount of training data. So they pre train local classifier at each node in hierarchical structure and then ensemble all local classifier into a global classifier using self-training. The entire process automatically determine the most appropriate level of each document. They have performed experiments on New York Times news articles. Corpus consists of 5 categories and 25 sub categories. The model that have been used is CNN. They have compared their results with other models and gives satisfactory results.\par
There are many existing approaches for document classification but few organize the document into the correct areas. This paper \cite{b7} proposed a new approach for hierarchical classification of text which is named as Hierarchical Deep learning for text classification. This paper employed new methods of deep learning. In neural networks, deep learning is very effective. Deep learning can manage supervised, unsupervised and semi supervised learning. In this paper, approaches to the hierarchy of the classification of documents advance the concept of deep neural learning network. The authors brought together various approaches to create hierarchical document classifiers. They have used three neural networks DNN, CNN and RNN. DNN is used for first level of classification of documents. Lets suppose if the first model output is medical, in next level it will train only all medical science documents. So, DNN is trained on first hierarchical level and in next level, each hierarchy of document is trained with specified domain. They have collected data of articles from Web of science. Results were compared with conventional text classification methods like  K-NN, decision tree, Naive Bayes and came to know that RNN outperforms all of them. CNN performs second best. The results shows that by using deep learning approaches, we can get good improvement in performance as compared to traditional methods. The HDLTex approach clearly shows variations of RNN at the top level and CNN or DNN at the lower level produced consistently good results.\par
A topical text classification is basic issue for many applications like emotion analysis, question answering, auto-tagging customer queries and categorizing articles. To handle large amount of labels, hierarchical text classification has been used as an effective way to organize text. Authors have proposed a new technique named as Hierarchically Regularized Deep graph Cnn \cite{b8}, which will solve the problem of document level topical classification and also address issues in topical classification which are long distance word dependency and nonconsecutive phrases.
First documents are converted into graphs that is centered on the word co-occurrence. Every document represents as a graph of vectors. For nodes of graph, pretrained vectors are used as an input. Word2vec is used in this scenario. After that they have generated sub graph of words and apply graph preprocessing on them. They have used both RNN and CNN models for text classification. They have proposed deep graph CNN approach for classification of text. Dependency between the labels can improve classification results, so that why they have used recursive regularization. They have used two data sets RCV1 and NY Times for experiments. Authors compared their experimental results with traditional hierarchical text classification models. The results shows that proposed approach have achieved good results than other models. Many applications arrange documents in the form hierarchical structure where classes are divided into categories and subcategories. In previous papers flat based methods were proposed. They only predict the categories of last level and ignore the hierarchical structure. To solve this problem, some work has been done in hierarchical category structure. Nevertheless, these literature focuses mainly on certain local areas or global structure as a whole ignoring dependencies between different levels of hierarchy.\par
In short, it’s a challenging task in which records are allocated to different categories and every category have its dependencies. The challenges are, connection among texts and hierarchy must be identified when interpreting the semantics of documents. Next challenge is that the hierarchical category structure has dependencies among different levels, which is mostly ignored by past research papers. Another challenge is that when we are assigning document to hierarchical categories, we should consider overall structure of hierarchical tree rather than just considering local region. 
This paper \cite{b9} have proposed a new approach named as recurrent neural network based on hierarchical attention to solve all above challenges. This framework automatically compiles a document with the most appropriate level of categories.
In this paper, authors have first work on the illustration of documents and hierarchical framework. They have applied embedding layer. This layer is used to computes text and the structure of their hierarchy. Word2vec is used for this task. After that, they have applied Bi LSTM to improve semantic text representation. After enhancing text representation and their structure, attention based recurrent layer is applied to represent the dependency for each category level from top to down. After that hybrid approach is designed for making predictions of the categories for each level. The experiments were conducted on academic exercises and repository of patent papers. Educational dataset is categorized into three levels hierarchy while repository of patent papers into four levels hierarchy. Experimental results shows that HARNN approach is effective.\par
Ivana \textit{et al} \cite{b10} have done hierarchical classification on Indonesian news articles. This research is conducted to examine whether hierarchical multi label classification could provide desirable results in classifying Indonesian news articles. So authors have proposed a hierarchical multi label classification model for Indonesian news articles. Three kinds of methods are available, which are global approach, flat approach and top down approach. To handle multi label classification, authors have used calibrated label ranking and binary relevance. Dataset is collected from news website and it is organized into two level hierarchy. The news articles have been split into 10 categories and 4 subcategories. For classification algorithms that were used are j48, Naïve bayes and SVM.\par
In this paper \cite{b11}, the author has present a study on Hierarchical text classification for Arabic language. Previously no research work is done on Hierarchical Arabic text classification, however many studies shows that work is done on flat Arabic text classification.  The method used for Hierarchical Arabic text classification is first order Markov chain. Transition probability matrix is created for each category when performing text classification task in Markov chain approach then for every document, a scoring procedure is carried out for all training set categories. In this study, data is taken from Alqabas newspaper. The documents are divided into 3 level hierarchy. First they preprocess the data. These kind of activities helps to get rid of useless data .TF IDF is used for preprocessing. Then they train the data by creating markov chain matrices. For each Arabic character a number is assigned. This process is called mapping. Probability matrix is created for each category. The writers have compared their proposed solution with LSI. Results demonstrate that top level performance is relatively good while it keeps on decreasing in below levels.\par
Deep neural networks are becoming more common task for classifying text because of their strong expressive capacity and less practical technical requirement. Although it is so attractive, models of neural text recognition suffer from the absence of trained data in many applications. In this paper \cite{b12}, author have proposed a hierarchical attention network for text classification of document. They have evaluated their model on large scale data. Datasets that they have used are reviews of imdb, yelp and amazon. They have compared their model with baselines methods like SVM, LSTM and word based CNN. Their model develops the document vector by combining relevant terms to the sentence vector and then the sentence vector to the document vector. The experimental results shows that their model perform better than the previous models.\par
Mao \textit{et al.} suggested a method for hierarchical text classification. Authors have formulate hierarchical text classification as Markov decision process \cite{b13}. In this process, at each phase, the agent examines the current situation and makes decisions. In this paper, authors suggested a framework in which system learns label assignment policy to decide when to place object and where to interrupt the assignment process. Their label assignment policy place each object to its label in a hierarchical structure. At start policy place an object on the root label. At every point, the policy determines which label object will be put next until stop action has been taken. Hierarchical label assignment policy examines the hierarchical structure of labels, both through training and analysis, in a precise way, which eliminates identity prejudice frequently seen in previous local and global strategies. Hierarchical label assignment policy is more versatile than other approaches in learning when to stop that support leaf node prediction. When exploring the labels hierarchy during training, Hierarchical label assignment policy performs better as compared to flat and local methods. Authors have conducted experiments on five publically available datasets. Two datasets are related to news category which are RCV1 and NYT and other two are related to protein functional catalogue and gene prediction and last dataset is from Yelp Dataset Challenge. The models that have been used for feature encoding are TextCNNN proposed by Kim \textit{et al.} \cite{b14}, Hierarchical Attention Network proposed by Yang \textit{et al.} \cite{b12} and bow-CNN proposed by Johnson \textit{et al.} \cite{b15}. They have compared their results with state of art hierarchical text classification methods and neural methods. The experiments results shows that their method surpassing traditional hierarchical text classification methods notably.\par
Recently many approaches have been proposed for chinese text classification. Chinese text classification is difficult as compared to other language like English, owing to the features of the Chinese text itself. To improve quality of chinese text classification, Liu \textit{et al.} \cite{b16} proposed a hierarchical model architecture which can extract rigorous context and information in a sequence manner from Chinese texts named as hierarchical comprehensive context modeling network. This approach is a combination of LSTM and temporal convolutional networks \cite{b17}. LSTM is used for  extracting context and sequence features of document. In this paper authors have used four datasets for experimentation. The results conducted on CRTEXT dataset shows that their model achieved the best results
as compared to other datasets. This paper \cite{b26} proposed a new approach for a large-scale multi-label text classification. First text is converted into a graph structure then attention mechanism is used to represent the full functionality of the text. They have also used  convolutional neural networks for feature extraction. The proposed method shows good results as compared to state-of-the-art methods. In this paper \cite{b27} authors have proposed a neural network approach which is based on BiLSTM and Hierarchical Attention layer. They have performed experiments on different news datasets. They have compared their model with CNN and BiLSTM attention network and achieved better results.
\section{Hierarchical Text Classification Background}
\begin{figure}[htbp]
\centerline{\includegraphics{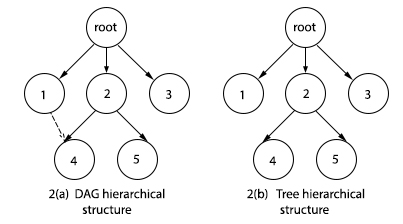}}
\caption{Illustration of hierarchical structure’s types}
\label{dag}
\end{figure}

Nowadays, text classification techniques are normally focused on a flat model. Currently, some methods for hierarchical text classification have been implemented to resolve the constraints of flat models and their methodology is focused on the classification method and feature selection. Mostly supervised learning technique is used which requires a category-based linear classification. In hierarchical classification, we are provided with different categories which are organized hierarchically. The corpus consist of training data in which documents are placed in categories and sub categories. Words or vocabulary are extracted from this corpus which have high frequency in a certain category. The vocabulary of every node is filtered and words are weighed according to their defined category. After this, test documents are evaluated and categories are ranked according to their term weighting. Term weights are assigned to each level node of hierarchy. These classifications allocate documents to categories on the basis of ranking. For categorizing, feature selection is used to determine categories of document. The words which are useless or have low frequency are removed to increase accuracy. There are different approaches that used to remove low frequency words.\\*
\begin{figure}[htbp]
\centerline{\includegraphics{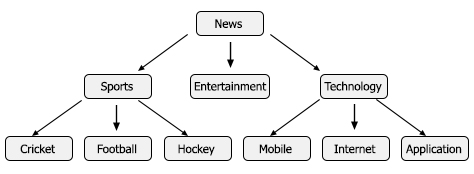}}
\caption{Example of Hierarchical Classification.}
\label{herar}
\end{figure}
For example in ``Fig.~\ref{herar}'', news is divided into three category i.e sports, entertainment and technology. On level 1, we can easily differentiate between all categories, if or not the document refers to sports, entertainment or technology. If text corresponds to sports, then we will further categories into sub categories, either the sports is cricket, football or hockey. In a hierarchy, each level represents the category of a document. A classifier is trained for each node in a hierarchical tree. When the new document arrives in a classification process, it is assigned to a suitable root node category then it use good category classifier to determine which direction to proceed in a hierarchy. And it will proceed until a document is assigned to a leaf node. There are two approaches for describing a hierarchical structure, which are tree structure and Direct Acyclic Graph also known as DAG. The difference between these two structures is that, in tree structure each node should have one parent while in Direct Acyclic Graph structure single node may have one or more parent. Fig.~\ref{dag} shows the illustration of both structure.
According to Freitas \textit{et al.} (2007) and Sun \textit{et al.} (2001), there are different types of hierarchical classification \cite{b1}, which are discussed below.
\subsection{Flat Classification}
\begin{figure}[htbp]
\centerline{\includegraphics{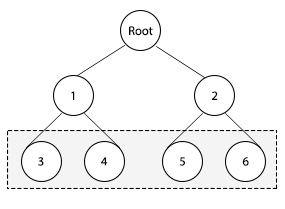}}
\caption{Flat classification approach.}
\label{flat}
\end{figure}

Koller and Sahami (1997) proposed this approach\cite{b2}. It is the most simple and straightforward method to handle hierarchical classification and it totally neglects class hierarchy. In flat classification you don't worry about parent categories, only just classify document to its final leaf node. This method works as a conventional algorithm for classification. Nonetheless, the problem of hierarchical classification is solved implicitly. If a leaf class is allocated to instance, almost all of its descendant classes are also allocated to it indirectly. The advantage of this method is its simplicity. It is an easy way to execute with an out of-the-box classifier. This approach also have some disadvantages. The main disadvantage is that this approach doesn't explore complete information of parent child class relationship which can lose important information and could decrease the efficiency. ``Fig.~\ref{flat}'' shows the illustration of flat classification approach. The dashed rectangle shows multi class classifier.
\subsection{Global (big bang) approach}
\begin{figure}[htbp]
\centerline{\includegraphics{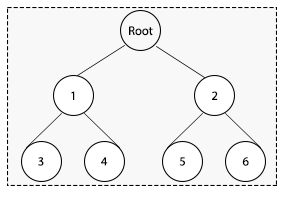}}
\caption{Global approach: one global classifier for the entire class hierarchy.}
\label{global}
\end{figure}

This approach is simple, relatively complicated model, in which the whole class hierarchy takes into account in a single run. That means it classify data in one run. Fig.~\ref{global} shows the illustration of one global classifier for the entire class hierarchy. When used in the test process, the induced model is identified by each test example, which can allocate classes to check example at any hierarchical level. This approach can go in different direction.  Many use clustering approach, some recreate the problem as a multi-label, and others have altered the latest algorithms. The advantage of this approach is that this approach have smaller model as compare to other models and the class relationship is determined during model development and it provide fast predictions. The drawback of this approach is that due to the high computations, there is higher chance of misclassify unseen data. However, due to the high complexity of this approach, it is rarely used.\\*
Labrou \textit{et al.} illustrate this approach. In their work \cite{b3}, the program classifies website as subcategory of Yahoo! hierarchical groups. This approach is unique to applications for text mining. A threshold is used for the final classification. There are some other method of global classifier which is based on multi label classification. Kiritchenko \textit{et al.} have done work on multi label classification\cite{b4}. To determine a hierarchical class, the learning process is amended to include all hierarchical classes by increasing the number of nodes having no child with information of predecessor classes.
\subsection{Hierarchically-Structured Local Classifiers}
The third approach is called as local approach which is also known as top down approach. This method uses predefined data classification scheme to construct a classification hierarchy. It uses hierarchy to develop classifiers utilizing local information. Benefits of this strategy are straightforward, effective and can handle multi-label hierarchical classification and it maintains natural information of data hierarchy. One major drawback of this approach is if there is classification mistake in parent node, it will be propagated downwards the hierarchy. It can greatly affect the result of classification. There are three methods to implement local classification approach named as local classifier per parent node, local classifier per node and local classifier per level.\\*
\subsubsection{Local classifier per parent node}
\begin{figure}[htbp]
\centerline{\includegraphics{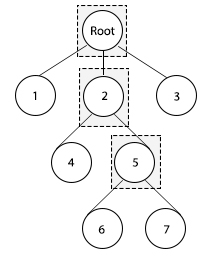}}
\caption{Illustration of local classifier per parent node (each dashed rectangle represents a multi-class classifier).}
\label{lcpn}
\end{figure}

This strategy is also called top down approach. In this approach, we train one multi class classifier for each parent node to differentiate between its child nodes. Now lets take an example of class tree as shown in Figure.~\ref{lcpn}  and approach that is used is class prediction of hierarchy. Lets assume document is assigned to class 2 node by initial classifier. Then in level 2 we have second classifier which is class 5 node, is learned with the children of class 2 node. Then class 5 node will make its class assignment. This process will continue until classifiers are available. In our example, we have one classifier on the first level and one classifier on second level. To stop incoherence with the prediction of different level, then you can develop a framework in which a document classified by first classifier as node 2, can only be seen by the node 2 classifier at the second level.
 \\*
\subsubsection{Local classifier per node}
\begin{figure}[htbp]
\centerline{\includegraphics{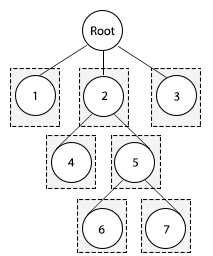}}
\caption{ Illustration of Local classifier per node approach (each dashed rectangle represents a binary classifier).}
\label{lcn}
\end{figure}

This popular is most popular which have been used in many researches. This method trains one binary classifier for each node of hierarchical class except root node. Figure.~\ref{lcn} shows that illustration of local classifier per node. There are a few issues with this strategy, first one is the hierarchical structure of local training sets is not taken into account. Secondly, it is restricted to issues where partial depth labeling examples involves. The benefit of this method is that it is by default multi label that means it can predict several labels per class hierarchy. \\*
\subsubsection{Local classifier per level}
\begin{figure}[htbp]
\centerline{\includegraphics{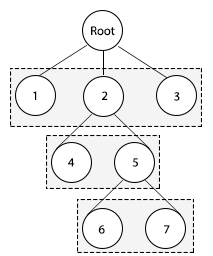}}
\caption{Local classifier per level (each dashed rectangle represents a multiclass classifier).}
\label{lcl}
\end{figure}

In this method we train one multi class classifier for each level of the class hierarchy. This classification method has not been widely used in researches. Figure.~\ref{lcl} shows that illustration of local classifier per level approach. Lets take this example, we have three classifier that will be trained to predict classes. The training data are built in the same manner as the parent node method is applied in the local classifier. The disadvantage of this approach is the inconsistency problem it presents.
\par
Local classifier approaches are very efficient. It uses hierarchical data information while preserving generality and smoothness. Nevertheless, the categorization method that have chosen, you could have a heavy final model in the end. In local classifier approach, there is also an issue of error propagation, which means if an error occur on first level, it will affect all the following ones down the hierarchy.
\section{Proposed Methodology and Dataset}
\begin{figure}[htbp]
\centerline{\includegraphics{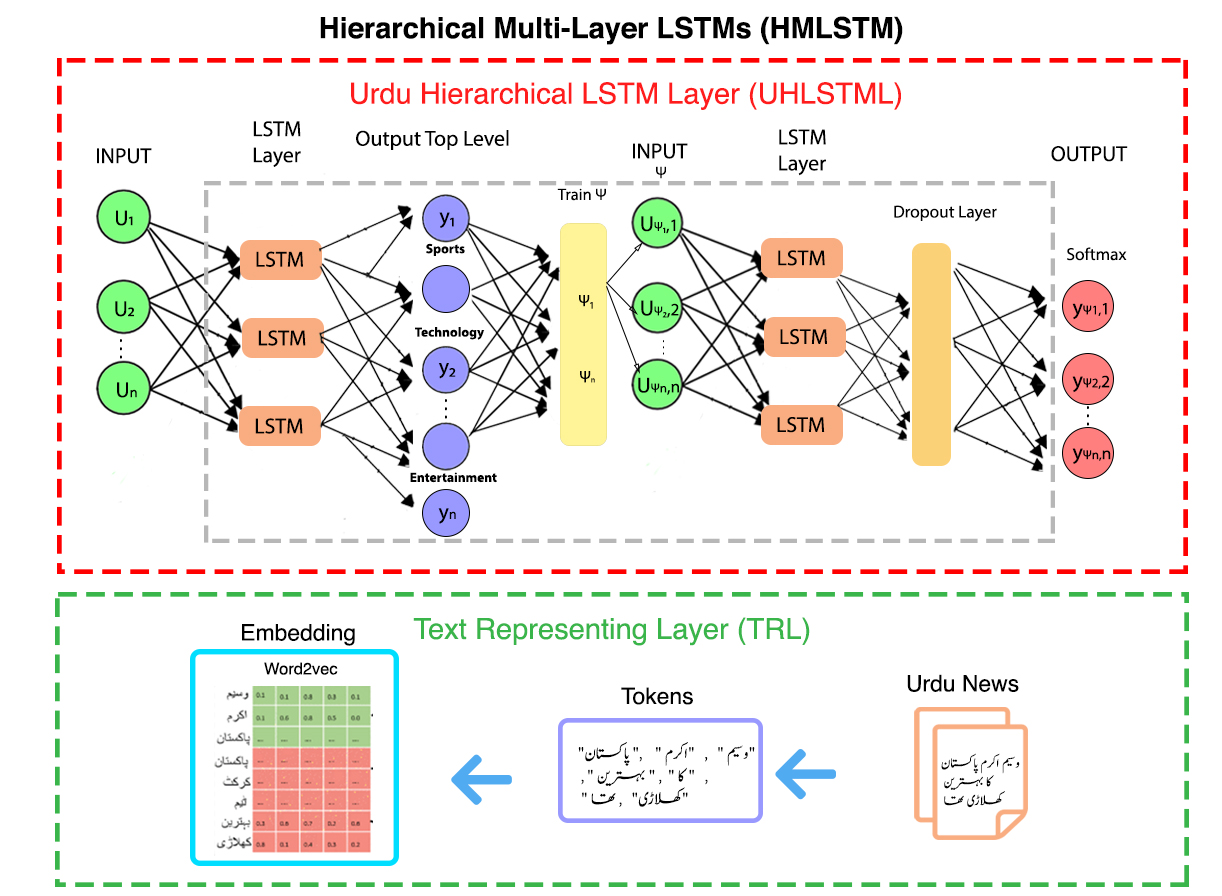}}
\caption{Hierarchical Multi-Layer LSTMs (HMLSTM)}
\label{full_approach}
\end{figure}

In this section, we will discuss our proposed methodology and self created dataset for hierarchical text classification of urdu news. As shown in figure ~\ref{full_approach}, our approach Hierarchical Multi-Layer LSTMs (HMLSTM) contains two parts that is Text Representing Layer (TRL) and Urdu Hierarchical LSTM Layer (UHLSTML). Mainly, we use the Text Representing Layer to achieve a unified representation of each urdu news text and the hierarchical category structure. After this, we apply our deep neural model Urdu Hierarchical LSTM Layer (UHLSTML) to predict hierarchical categories and sub categories of urdu news text.
\subsection{Text Representing Layer}
In first phase of our approach HMLSTM, text representing layer TRL purpose is to make a unified representation of each urdu news text and their hierarchy structure. As we know, it is complicated for machine learning algorithms to use natural language directly. So we use embedding for this, which helps to convert words into a numeric representation. In TRL, an embedding layer is used to encode text. We first convert urdu news text U into tokens and take hierarchical category labels L as an input. After this, we apply word2vec \cite{b21} embedding, word2vec requires list of tokens of a document then start vector encoding of tokens. The fundamental principle of Word2vec is that rather than representing words as one-hot encoding i.e tf-idf vectorizer in high dimensional space, we define words in dense low dimensional space in such a way that identical words get identical word vectors, such that they are projected to adjacent points. It learns to represent words with similar meanings using similar vectors. Word2vec accepts text corpus as an input and converts text into numerical data which will later pass it to our model as an input.
We can formulize urdu news text U as a sequence of N words \begin{math} U=\left(v_{1},v_{2},\ldots,v_{N}\right) \end{math} where \begin{math} u_{i}\in \operatorname R_{h} \end{math} is initialized by h-dimensional of the feature vectors with Word2vec .\par
As our approach is supervised learning we have labeled data in the form of categories and sub categories. Our proposed model UHLSTML take hierarchical category structure labels L as an input. We can formulize labels L as a sequence of 12 classes. \begin{math} L=\left(l_{1},l_{2},\ldots,l_{12}\right) \end{math}. So similar to word2vec, we have to convert categorical data into numerical data. There are several methods to transform categorical values to numerical values. Every method has certain effects on the feature set. Here we are using label encoding for categories and sub categories labels. As we have 12 classes, 3 categories and 9 sub categories. Label encoding convert these labels in machine readable form. So, Label L is an input vector which has integers representing various classes in the data. As already discussed, UHLSTML is deep neural model, it works well if we have classes distributed in a binary matrix. So, we convert label class vectors (ordinal numbers) to binary matrix so that our labels are easily understandable by UHLSTML. For this, we have used tensorflow keras \cite{b22} to-categorical which converts a class vector (integers) to a matrix of binary values (either ‘1’ or ‘0’).
After this our text representation layer obtained whole representation of data and achieved a unified representation of each urdu news text and the hierarchical category structured labels, we pass it towards our propose model UHLSTML.
\subsection{Urdu Hierarchical LSTM Layer}
\begin{figure*}[htbp]
\centerline{\includegraphics{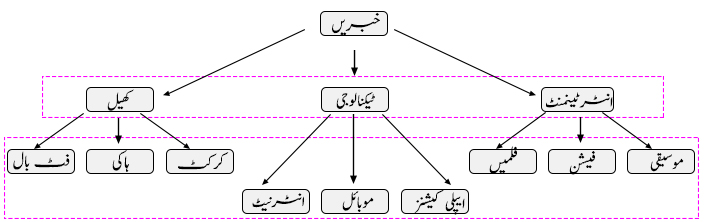}}
\caption{LSTM layer per level (each dashed rectangle represents a LSTM layer).}
\label{herar_urdu}
\end{figure*}

Urdu Hierarchical LSTM layer is based on Hierarchically-Structured Local Classifiers approach in which we have Local classifier per level method. In this method we train one LSTM layer for each level of the class hierarchy as shown in figure ~\ref{herar_urdu}. \par
\begin{figure}[htbp]
\centerline{\includegraphics{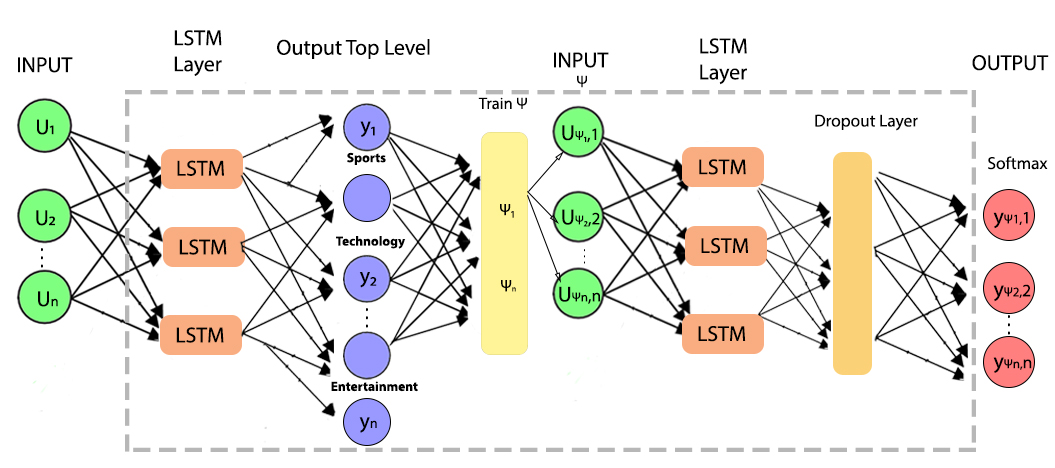}}
\caption{The proposed Urdu Hierarchical LSTM Layer having two LSTM layers  and  one dropout layer.}
\label{approach}
\end{figure}

After receiving combined and structured representation of urdu news text and labels, we propose UHLSTML, an  end-to-end  fully  connected deep LSTMs network to perform automatic feature learning. It is based on recurrent neural networks architecture. Figure~\ref{approach} shows the architecture of the proposed network, which has two LSTM layers, one LSTM layer for top categories and second LSTM layer for sub categories and a softmax layer that gives the predictions. First layer will also be used as feedforward layer, it learns the features of top level category that is sports, technology and entertainment and forward all these features to second LSTM layer.\par
The input layer contains text features. We pass input which is urdu news text U to first LSTM layer which learns features of top category and gives output of y. First LSTM layer is used for first level of classification. After this, we pass these features as an input to second LSTM layer. The second level classification in the hierarchy consists of a trained top category. In second LSTM layer, each second level hierarchy is connected to the output of the first level. Let's take an example in this~\ref{approach} case, if the output of our first LSTM is labeled as Sports then the second LSTM is trained only with all sports related news like football, hockey and cricket. Which means first level LSTM is trained with all news whereas second level LSTM will only trained with the news of specific category.
The LSTMs in this research are trained with a standardized back propagation algorithm that uses ReLU as an activation function and a softmax layer that gives the output. Optimizer that we used is Adam having learning rate 0.001. To calculate loss, categorical crossentropy is used. LSTM dropout is applied to the last LSTM layer to reduce over fitting and allow extra efficient learning. \par
The LSTM network consists of various memory blocks which are called cells. LSTM module composed of 3 gates named as Forget gate, Input gate and Output gate. In Forget Gate, it determine which details should be removed from the cell in that specific time duration. Information which is no longer needed for the LSTM to comprehend items or information that is of less value is eliminated. This is handled by sigmoid function. Forget gate takes in two inputs, previous state ht-1 and input at that particular time x-t and outputs a vector, with values ranging from 0 to 1. This vector value is multiplied to the cell state.
\begin{equation}
f_{t}=\sigma\left(W_{f} \cdot\left[h_{t-1}, x_{t}\right]+b_{f}\right)
\end{equation}
In input gate, we add some new information to the cell state. Sigmoid function decides, which value is added to the cell state. The values which are passed tanh function gives weightage to them which results values from -1 to +1.
\begin{equation}
\begin{aligned}
i_{t} &=\sigma\left(W_{i} \cdot\left[h_{t-1}, x_{t}\right]+b_{i}\right) \\
\tilde{C}{t} &=\tanh \left(W{C} \cdot\left[h_{t-1}, x_{t}\right]+b_{C}\right)
\end{aligned}
\end{equation}
In Output Gate, the role is to pick valuable information from the current cell and makes it to the output through the output gate. It is handled by Sigmoid function, which value is added to the cell state and tanh function assigns weights to the values and multiplying the value of this to the output of Sigmoid. Here is an equation.
\begin{equation}
\begin{array}{l}
o_{t}=\sigma\left(W_{o}\left[h_{t-1}, x_{t}\right]+b_{o}\right) \\
h_{t}=o_{t} * \tanh \left(C_{t}\right)
\end{array}
\end{equation}
\subsection{Proposed Dataset}
\begin{figure}[htbp]
\centerline{\includegraphics{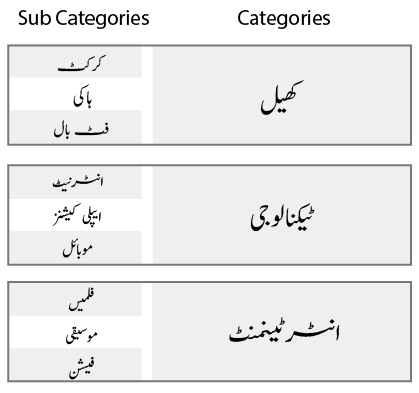}}
\caption{Urdu-News-Dataset-for-Hierarchical-Text-Classification}
\label{undhtc}
\end{figure}

In supervised learning, the performance of the model mainly relies on dataset. The learning of a neural network relies on the dataset and if there is less amount of dataset, the learning will be insufficient and it will decrease the accuracy of model. There are alot of publicly available dataset like Reuters-21578, 20 Newsgroups and RCV1 for english news but there is limited amount of large dataset for urdu news. So large datasets are required to perform various natural language tasks and applications. To provide a suitable dataset for model training and evaluation of model results, we created a urdu news corpus which is named as Urdu News Dataset for Hierarchical Text Classification (UNDHTC) which will be publicly available at github \footnote{https://github.com/TaimoorAhmedJaved/Urdu-News-Dataset-for-Hierarchical-Text-Classification}. The main goal of corpus development is to make use of it in comparison and evaluation in hierarchical classification and to help researchers in future to perform NLP tasks. UNDHTC is classified into 12 categories that is 3 categories Sports, Technology and Entertainment which is further classified into 9 sub categories cricket, hockey, football, applications, mobile, internet, music, fashion and movies as shown in figure ~\ref{undhtc}.
\subsubsection{Dataset Creation}
Most of data was collected from popular Pakistani urdu news websites which are Daily Pakistan\footnote{(www.dailypakistan.com.pk}, Urdupoint\footnote{www.urdupoint.com}, Express news\footnote{(www.express.pk}, Dawn news\footnote{www.dawnnews.pk}, BBC Urdu\footnote{(www.bbc.com/urdu}, ARYnews\footnote{www.urdu.arynews.tv}, BOL news\footnote{(www.bolnews.com/urdu} and Hum news\footnote{www.humnews.pk}. To  automate  collection of data, we  have used Selenium with Chrome driver for Chrome web browser.
\subsubsection{Annotation}
The dataset labels are in twelve categories. We labeled our dataset with the support of our coworkers who have a solid knowledge of Urdu language. The proper instructions and exemplars provided them for annotation of data. The labels were assigned according to the context and subject of the sentence.
\subsubsection{Dataset Statics}
Approximately, 299815 news were extracted from different websites by using scraping tools. 57566 news were cleaned and labeled as shown in table ~\ref{Statics} and total number of documents in categories are shown in table ~\ref{detail}.
\begin{table}[ht]
\centering
  \caption{ Dataset Statics}
  \label{Statics}
  \begin{tabular}{lll}
    \toprule
     \multicolumn{2}{c}{\textbf{Dataset Statistics}}\\ 
    \midrule
     Total News        & 299815         \\  
     Cleaned        & 57566         \\  
     Labeled  &  51325\\
    \bottomrule
  \end{tabular}
\end{table}
\begin{table}[ht]
\centering
  \caption{Dataset Stats in Detail}
  \label{detail}
  \begin{tabular}{llll}
    \toprule
\textbf{Category} & \textbf{Sub Category} &\textbf{Number of documents} & \textbf{Total}\\
    \midrule
                      & Cricket   & 12431 \\
           Sports         & Hockey & 2011 & 20002\\
                     & Football   & 5560 \\
    \midrule
                      & Internet   & 1735 \\
    Technology        & Applications & 3742 & 16208\\
                     & Mobile   & 10731 \\      
    \midrule
                      & Movies   & 3902 \\
    Entertainment     & Music & 2552 & 15115\\
                      & Fashion & 8661 \\
    \bottomrule
  \end{tabular}
\end{table}
In next section, we will discuss experiments that were performed on this dataset and also compare results with baseline methods.
\section{Experiments}
\begin{figure}[htbp]
\centerline{\includegraphics[width=0.79\textwidth]{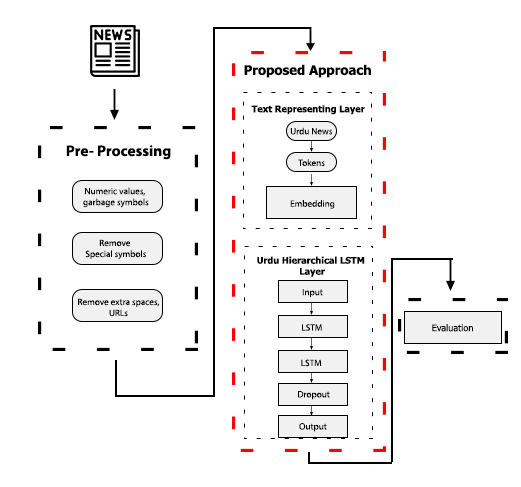}}
\caption{Experimental Flow Diagram.}
\label{flow}
\end{figure}

In this section, we first introduce the dataset and apply preprocessing steps. Then we conduct extensive experiments and compare our proposed approach with state-of-the-art traditional machine learning algorithms and deep neural model. In the end we will discuss experimental results. Figure~\ref{flow} shows complete flow diagram of our experiment.

The total number of 57566 news were used in this experiment. UNDHTC dataset is manually labeled which are divided into total 12 categories which contain 3 top categories and 9 sub categories. The training/test split for UNDHTC is done by ourselves, which is 80\% for training and 20\% for test as shown in table ~\ref{test/split}. We randomly sample 20\% from the training data .
\begin{table}[ht]
\centering
  \caption{ The training/test split for UNDHTC}
  \label{test/split}
  \begin{tabular}{llll}
    \toprule
\textbf{Dataset} & \textbf{Training}&\textbf{Testing}&\textbf{Class/Labels}  \\
    \midrule
51325 & 41060 & 10265 & 12  \\
    \bottomrule
  \end{tabular}
\end{table}
\subsection{Preprocessing}
We have applied basic pre-processing methods on our data. Data pre-processing in every language is an essential step taken before applying any machine learning algorithm or deep neural network. It helps to remove unnecessary data in the corpus. English characters, special symbols, numeric values, and URLs are omitted such that the text includes only the target Urdu language.  The structured representation of text data is efficiently controlled to improve the experiment’s accuracy and efficiency. Preprocessing steps that we applied on our dataset are as follows.
\subsubsection{Tokenization}
\begin{figure}
\centerline{\includegraphics[width=0.3\textwidth]{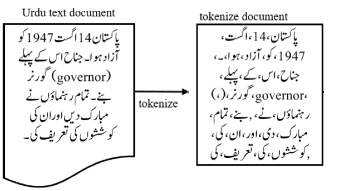}}
\caption{Urdu tokenize document}
\label{token}
\end{figure}

This is the initial step in the processing of any language of the NLP task. It transforms the phrases into valuable tokens or a single expression. It transforms raw text into a tokens chart, in which every token is a Word. Inaccurate tokenization can affect the results of the experiment. Urdu is a less linguistic tool and its signs face different problems as spatial incoherence between terms and sentence limit detection. Here is a figure ~\ref{token} which show urdu text converted into tokens.
\subsubsection{Stop Words Removal}
To remove unnecessary words we use stop words. It is used for removing words that carry little to no meaning. Natural languages are consist of meaningless words and functional words. In Urdu, we call stopwords as Haroof-e-Jar and in English they are called conjunctions. Stop words are often omitted from the corpus before NLP is done. By removing these words, we can get better results in our research work.
\subsubsection{Punctuation Removing}
In Urdu, mostly used punctuation are ‘-‘, ’\_’, and ‘.’. Such symbols do not have much value to be used as a feature for classification tasks. So we remove these symbols by using regular expressions.
\subsection{Experimental Setup}
All of our experiments were run on 6 core Intel Core i7 9750 @2.60GHz, with 16 GB RAM and GTX 1660ti GPU. The software that we used is jupyter notebook having Python 3.7.6. In our experiments, we used public version of word2vec in which training algorithm CBOW is used to train 100 dimensional word embeddings with  minimum count of words 5, filter window size is 5 and other default parameters. For deep learning model, the parameters of training the models were set as batch size 32, drop out 0.5 and other default parameters. Data was trained on 80\% for training and 20\% randomly selected for test as shown in table ~\ref{test/split}. We have done our hardest to use the best parameters in our experiments.
\subsection{Baselines and Deep Neural Model Comparison }
We compare our proposed approach HMLSTM with several baseline methods, such as Naïve Bayes, Support Vector Machines (SVMs), Decision Tree, Random Forest, K-Nearest Neighbors, Logistic Regression and deep neural model such as convolutional neural network (CNN).
\begin{itemize}
\item \textbf{Gaussian Naïve Bayes} :
We have used global approach in this method that builds a single decision tree to classify all categories simultaneously. In this method, we first convert urdu news text in the form of words. After this we use one hot encoding to encode those words.  We encode both columns category and sub category having labels by using label encoding. Then we combine both the labels (category and subcategory) in a single variable. After that, we split the dataset to the train and the test parts. Then we apply GaussianNB algorithm using binary relevance with default parameters in which variance is 1.0 raise to power -9 and perform predictions. skmultilearn library is used in it \cite{b24}.
\item \textbf{Support Vector Machines (SVMs)}:
In this method, we perform standard multi-label classification using one-vs-the-rest (OvR) strategy because we have multiple labels, so we use it. Scikit-learn is a pipeline framework for automating workflows for training. In machine learning applications, pipelines are very common because there are many data to manipulate and many transformations to be carried out. So we used pipeline to train OnevsRest classifier. We use LinearSVC with OneVsRestClassifier and other default parameters.
\item \textbf{Decision Tree}:
In this method, we also used global approach strategy. Data Reading is done using Pandas. We first convert urdu news text in the form of tokens. After this we use word2vec embedding to encode those words. For labels, we use label encoding. Label encoding convert these labels into vectors. As, we have multiple labels so we used multilabel classification. Multilabel classification tasks can be handled using decision trees. In Decision Tree Classifier, we set criterion='entropy' and remaining parameters are set as default and perform predictions. sklearn library \cite{b25} is used for classification.
\item \textbf{Random Forest}:
Random forest is also used for multilabel classification task. So we used this method in hierarchical text classification. Random Forest can inherently deal with multiclass datasets. As a result, we adopted a one-versus-rest strategy using random forest classifier. Regular Random Forest cannot perform well on multilabel classification \cite{b23}. Experimental results shows better classification performances with one versus all classifier. In random forest, we set parameters as default and perform predictions.
\item \textbf{K-Nearest Neighbors}:
In this method, we first convert data and labels into vectors by using embedding and label encoding. After that, we split the dataset to the train and the test parts then apply k-nearest neighbors (KNN) algorithm. We set number of neighbors 3 with other default parameters and perform predictions.
\item \textbf{Logistic Regression}:
In this method, we also perform standard multi-label classification using one-vs-the-rest (OvR) strategy. Scikit-learn \cite{b25} is a pipeline framework for automating workflows for training. Here we also used pipeline to train OnevsRest classifier and apply logistic regression on data. L2 regularization is applied in logistic regression with other default parameters and perform predictions.

\item \textbf{Convolutional Neural Network}:
In CNN, we first convert urdu news text in the form of tokens. After this we use word2vec embedding to encode those words. Then we apply CNN on our dataset, we pass the kernel over these embeddings to find convolutions. CNN model is trained with two layers, one is convolution and other is softmax layer. We also applied “relu” activation layer on the output of every neuron. Drop out layer is also applied to reduce overfitting.
\end{itemize}
\subsection{Evaluation Metrics}
To evaluate the performance of algorithms we have used three common evaluation metrics i.e. Precision, Recall, and F1 score which includes Micro-F1 and Macro-F1. We also used Classification Accuracy, which is also known as term accuracy. Let \begin{math}
TP_{l}, FP_{l} , FN_{l} \end{math} be the number of true positives, false positives and false negatives for the l-th label in the category C respectively. Then the equation of Micro-F1 is
\begin{equation}
P=\frac{\sum_{l \in \mathcal{C}} T P_{l}}{\sum_{l \in \mathcal{C}} T P_{l}+F P_{l}} , R=\frac{\sum_{l \in \mathcal{C}} T P_{l}}{\sum_{l \in \mathcal{C}} T P_{l}+F N_{l}}, F_{1 Micro}=\frac{2 P R}{P+R} \\
\end{equation}
Equation for Macro-F1 is defined as
\begin{equation}
P_{l}=\frac{T P_{l}}{T P_{l}+F P_{l}}, R_{l}=\frac{T P_{l}}{T P_{l}+F N_{l}}, F_{1 Macro}=\frac{1}{|\mathcal{C}|} \sum_{l \in \mathcal{C}} \frac{2 P_{l} R_{l}}{P_{l}+R_{l}}\\*
\end{equation}
\subsection{Qualitative Analysis}
\begin{figure}[htbp]
 \centerline{\includegraphics[width=0.85\textwidth]{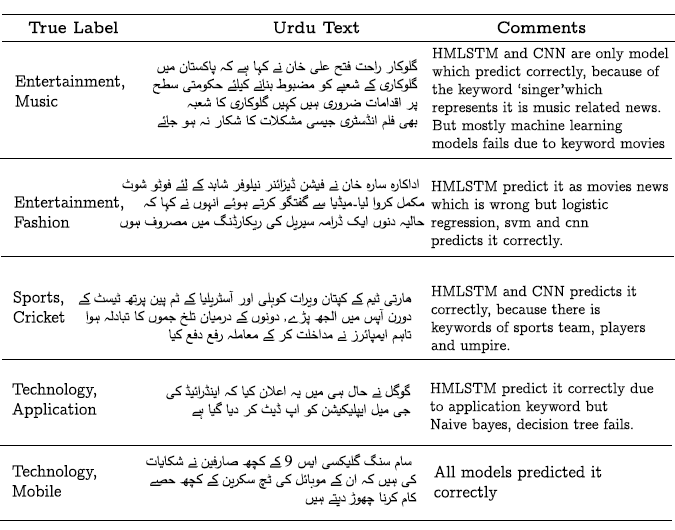}}
     \caption{Qualitative Analysis of HMLSTM results and other baseline methods}
       \label{qualitative}
\end{figure}

To measure the performance of algorithms across different examples, we summarizes some of the examples from evaluation set as shown in figure~\ref{qualitative}.We can see that HMLSTM works well in most of the cases because LSTM can look back to many time steps which means it retain long term memory when dealing with long sentences. In first example, we can see that all baseline approaches fails, but HMLSTM and CNN predict correctly. However our model still needs improvement, as we can see in second example our model fails because it didn't consider context of news due to presence of keywords like actor and drama.
\subsection{Experimental Results}

\begin{table*}
\centering
\caption{HMLSTM and other model Results on Urdu News Dataset}
\label{results}
 \begin{tabular}{lllllll}
\toprule
Methods    & \#Classes & Accuracy & Precision & Recall & Macro-F1 & Micro-F1  \\ 
\midrule
 Gaussian Naïve Bayes & 12 & 0.4744 & 0.4223 & 0.8693 & 0.5257 & 0.6987 \\
 Support Vector Machines & 12 & 0.7711 & 0.7434 & 0.7769 & 0.7590 & 0.8878 \\ 
 Decision Tree    & 12 & 0.7285 & 0.6113 & 0.6092 & 0.6102 & 0.7949\\ 
 Random Forest    & 12 & 0.7959 & 0.9256 & 0.6331 & 0.7391 & 0.8806 \\ 
 K-Nearest Neighbors & 12 & 0.5989 & 0.6300 & 0.5668 & 0.5467 & 0.6940 \\
 Logistic Regression    & 12 & 0.7962 & 0.7875 & 0.7822 & 0.7848 & 0.8980\\
 CNN    & 12 & 0.8740 & 0.8866 & 0.8496 & 0.8632 & 0.9323 \\
 HMLSTM    & 12 & \textbf{0.9402} & \textbf{0.9259} & \textbf{0.8812} & \textbf{0.8927} & \textbf{0.9683}\\
\bottomrule
\end{tabular}
\end{table*}
To demonstrate practical significance of our proposed model, we compare HMLSTM with all the baselines. The experiments perform on urdu news dataset (UNDHTC) shows that our proposed approach Hierarchical Multi-Layer LSTMs (HMLSTM) outperforms the machine learning models. The table~\ref{results} shows comprehensive results of all models. CNN performed well and gave promising result as compare to state-of-the-art methods. For traditional text classification algorithms, we can see that Logistic Regression using one-vs-the-rest (OvR) strategy and applying regularization, it outperform other models. Then we have Support Vector Machines after LR, which is performing better than other four algorithms. SVM is good in performing multi label classification, that is why its better than rest. While K nearest neighbor perform least on urdu news dataset.\\
\begin{figure}[htbp]
 \centerline{\includegraphics[width=\linewidth]{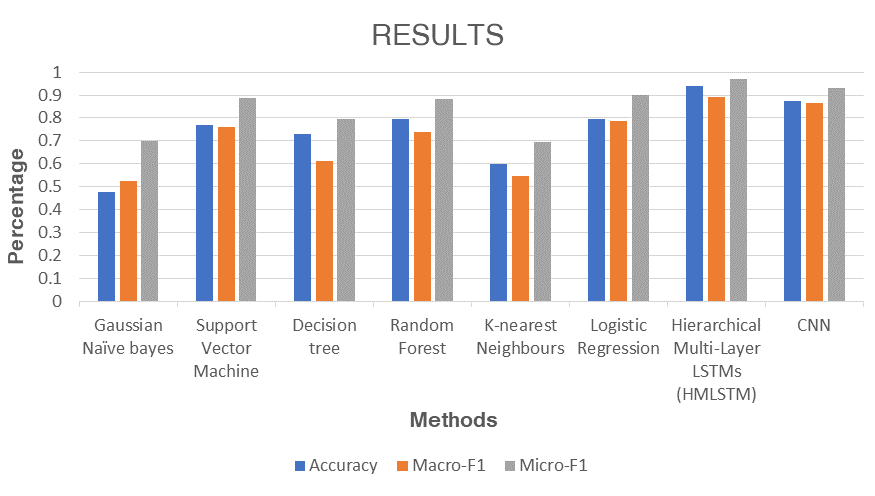}}
     \caption{Results showing Accuracy, Macro-F1 and Micro-F1}
       \label{chart}
\end{figure}
By splitting labels according to their levels, we evaluate causes of performance gains. Table~\ref{results} indicates Macro-F1 variations between the base models. 
HMLSTM achieves the best results in all evaluation metrics, which shows HMLSTM is more competent for doing hierarchical multi label text classification tasks with hierarchical group structure and it deal with hierarchical category adequately and correctly. The main reason is that HMLSTM could identify the correlations between the Urdu news texts and labels, while other algorithms can't. HMLSTM integrate the predictions of every level in the form of hierarchical structure.
\begin{figure}[htbp]
 \centerline{\includegraphics[width=\linewidth]{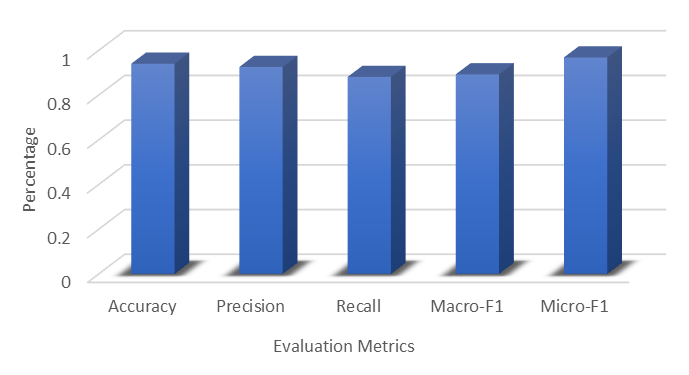}}
     \caption{Hierarchical Multi-Layer LSTMs (HMLSTM) Results}
       \label{approach_results}
\end{figure}
The bar chart as shown in figure~\ref{approach_results} clearly demonstrates the effectiveness of our proposed approach for hierarchical text classification.

\section{CONCLUSION AND FUTURE WORK}
In this paper, we have proposed Hierarchical Multi-Layer LSTMs method built upon deep neural network which is used for hierarchical text classification of urdu news text. We first applied Text Representing Layer for obtaining numeric representation of text and categories by applying embedding. After receiving combined and structured representation of urdu news text and labels, we applied Urdu Hierarchical LSTM Layer, an end-to-end fully connected deep LSTMs network to perform automatic feature learning. Finally, extensive experiments is performed on our self created dataset named as Urdu news dataset for hierarchical text classification. Our experimental results clearly demonstrate that our method outperforms baseline methods significantly and also performed well as compare to neural model CNN. In the future, we are interested to concentrate further on improving the network structure in order to further improve the efficiency of our method. We will also try to develop Bi-LSTM based approach to to obtain improved hierarchical structure. In future, we will also try to optimize our model because it takes long training time.

\bibliographystyle{ACM-Reference-Format}
\bibliography{sample-base}


\end{document}